\documentclass{article} 
\usepackage{iclr2025_conference,times}
\usepackage{graphicx}

\usepackage{amsmath,amsfonts,bm}

\def\eqref#1{equation~\ref{#1}}

\def\1{\bm{1}}

\DeclareMathAlphabet{\mathsfit}{\encodingdefault}{\sfdefault}{m}{sl}
\SetMathAlphabet{\mathsfit}{bold}{\encodingdefault}{\sfdefault}{bx}{n}

\usepackage{hyperref}
\usepackage{url}
\usepackage{natbib}
\title{Shortcut Learning Susceptibility \\in \\Vision Classifiers}

\author{Pirzada Suhail\thanks{Corresponding Author} ,  Vrinda Goel \& Amit Sethi \\
IIT Bombay\\
\texttt{\{psuhail,vrinda.goel,asethi\}@iitb.ac.in} \\
}

\iclrfinalcopy 
\begin{document}

\maketitle

\begin{abstract}
Shortcut learning, where machine learning models exploit spurious correlations in data instead of capturing meaningful features, poses a significant challenge to building robust and generalizable models. This phenomenon is prevalent across various machine learning applications, including vision, natural language processing, and speech recognition, where models may find unintended cues that minimize training loss but fail to capture the underlying structure of the data. Vision classifiers based on Convolutional Neural Networks (CNNs), Multi-Layer Perceptrons (MLPs), and Vision Transformers (ViTs) leverage distinct architectural principles to process spatial and structural information, making them differently susceptible to shortcut learning. In this study, we systematically evaluate these architectures by introducing deliberate shortcuts into the dataset that are correlated with class labels both positionally and via intensity, creating a controlled setup to assess whether models rely on these artificial cues or learn actual distinguishing features. We perform both quantitative evaluation by training on the shortcut-modified dataset and testing on two different test sets—one containing the same shortcuts and another without them—to determine the extent of reliance on shortcuts. Additionally, qualitative evaluation is performed using network inversion-based reconstruction techniques to analyze what the models internalize in their weights, aiming to reconstruct the training data as perceived by the classifiers. Further, we evaluate susceptibility to shortcut learning across different learning rates. Our analysis reveals that CNNs at lower learning rates tend to be more reserved against entirely picking up shortcut features, while ViTs, particularly those without positional encodings, almost entirely ignore the distinctive image features in the presence of shortcuts.
\end{abstract}

\section{Introduction}

Machine learning models are expected to learn meaningful patterns from data to make accurate predictions and generalize well across different domains. However, in many cases, models do not learn the intended task-relevant features but instead rely on shortcut learning, where they exploit spurious correlations in the training data that happen to be predictive of the labels. Rather than learning intended patterns, models often latch onto unintended statistical correlations present in the training data, leading to poor generalization across different domains. While shortcut learning may improve performance on in-distribution test data, it significantly degrades model robustness when evaluated on out-of-distribution samples in a real-world setting where these spurious correlations no longer hold.

In computer vision, different classification architectures process input data in fundamentally distinct ways, which may lead to varying tendencies toward shortcut learning. Since each of these architectures encodes spatial and structural information differently, their responses to shortcuts may vary. However, the specific nature of their susceptibility to shortcut learning remains unclear. In this paper, we aim to systematically evaluate how CNNs, MLPs, and ViTs learn from data when deliberate shortcuts are introduced into the training dataset. Specifically, we modify certain pixel regions in the images to correlate deterministically with class labels, embedding artificial cues that serve as an unintended yet easily learnable path to minimize training loss. These modifications act as an alternative path for models to reduce training error, creating a controlled setup where we can analyze whether models rely on these shortcuts or learn the actual distinguishing features in the images.

To measure the extent to which models depend on shortcut cues, we train CNNs, MLPs, and ViTs on the modified datasets and evaluate them on two test sets: one containing the same shortcuts and another without any shortcuts. If a model primarily learns the artificial shortcut instead of the actual class-relevant features, its performance is expected to drop significantly when evaluated on the test set without shortcuts. Beyond performance metrics, qualitative evaluations is performed using network inversion-based reconstruction techniques to analyze what the models internalize in their weights. We also explore how learning rate variations impact shortcut reliance across different architectures.

\section{Related Works}

Shortcut learning is a phenomenon where machine learning models prioritize learning simple, potentially misleading cues from data that do not generalize well beyond the training set. This issue has been identified as a fundamental limitation of deep learning models, leading to poor robustness and transferability \cite{Geirhos_2020}. The study in \cite{hermann2024on} examines shortcut learning from a theoretical perspective by investigating the factors that influence whether a model will rely on a shortcut. This phenomenon is particularly concerning in medical AI, where it has been linked to algorithmic unfairness. In \cite{Brown_2023}, the authors propose a method to detect shortcut learning in clinical machine learning models by applying multitask learning to identify improper correlations that may cause biased predictions. Beyond classification, the phenomenon has also been studied in medical image segmentation, where commonly used dataset preparation techniques, such as zero-padding and center-cropping, introduce unintended shortcuts that influence segmentation accuracy\cite{10.1007/978-3-031-72111-3_59}. The paper \cite{bleeker2024demonstrating} investigates shortcut learning in vision-language models and evaluates how contrastive learning-based models tend to latch onto unintended patterns in multi-caption training scenarios. Shortcut learning has also been studied in the context of Vision Transformers. The work in \cite{10250856} explores how ViTs might be particularly prone to shortcut learning due to their reliance on self-attention mechanisms. The authors introduce a saliency-guided ViT model that leverages computational visual saliency maps to guide ViTs toward learning meaningful features rather than background artifacts.

This paper exploits network inversion-based reconstruction techniques to analyze different model architectures for their shortcut learning susceptibility. Inversion has been studied in \cite{KINDERMANN1990277,784232} using the back-propagation and evolutionary algorithms for feed-forward networks that identify multiple inversion points simultaneously, providing a more comprehensive view of the network’s input-output relationships. Later inversion of Convolutional Neural Nets was performed in \cite{suhail2024networkcnn}, using a conditioned generator that learns the input space of the trained models. Recent work by \cite{liu2022landscapelearningneuralnetwork} proposes learning a loss landscape where gradient descent becomes efficient, significantly improving the speed and stability of the inversion process. Later, \cite{suhail2024network} proposes an alternate approach to inversion by encoding the network into a Conjunctive Normal Form (CNF) propositional formula and using SAT solvers and samplers to find satisfying assignments for the constrained CNF formula. Also, \cite{ansari2022autoinverseuncertaintyawareinversion} introduces an automated method for inversion by seeking inverse solutions near reliable data points that are sampled from the forward process and used for training the surrogate model.

We are specifically interested in trying to reconstruct the training data as perceived by the models in presence of shortcuts. The work in \cite{haim2022reconstructingtrainingdatatrained} explores the extent to which neural networks memorize training data, revealing that a significant portion of the training data can be reconstructed from the parameters of a trained neural network classifier. Later, \cite{buzaglo2023reconstructingtrainingdatamulticlass} improves on these results by showing that training data reconstruction is not only possible in the multi-class setting, but that the quality of the reconstructed samples is even higher than in the binary case. In this paper we use network inversion-based reconstruction method proposed in \cite{suhail2024net} to understand the internal representations of neural networks and the patterns they memorize during training and compare the shortcut learning susceptibility of different vision classifier architectures.

\section{Methodology}
Our approach to analyzing shortcut learning susceptibility in different vision classifier architectures involves introducing two types of deliberate shortcuts into the dataset, including (1) \textbf{Positional shortcuts}, where specific pixel regions are modified such that their locations deterministically correlate with the class labels, and (2) \textbf{Intensity-based shortcuts}, where the pixel intensity in a certain region of the image is altered in a way that correlates with the class labels.

We analyse the shortcut learning susceptibility of classifiers trained on three distinct architectures including MLPs, CNNs, and ViTs.  These architectures inherently differ in how they process spatial and structural information, which affects their susceptibility to shortcut learning. The models are trained on the shortcut-modified dataset, and their generalization capabilities are evaluated by comparing their performance on two test sets: (1)\textbf{ one with the same shortcut modifications} and (2) \textbf{another without shortcuts}. A model that generalizes well and does not rely on shortcuts should perform comparably on both test sets, whereas a model that latches onto shortcuts will show significant degradation in performance on the latter. To quantitatively evaluate shortcut susceptibility, we define \textbf{Accuracy Difference} as the absolute difference between the model's accuracy on the shortcut test set and the normal test set, indicating the extent to which the model relies on shortcut features for classification, providing a measure of the inconsistency in model confidence across the two test sets.

Further, qualitative evaluation, performed using network inversion-based reconstruction technique aims to generate training-like samples based on the classifier's learned representations using a composite loss function as defined in \cite{suhail2024net} that integrates multiple constraints ensuring that the reconstructed data closely resembles the perceived training distribution.

\section{Experiments and Results}
In this section, we evaluate the shortcut learning susceptibility of three different vision classifier architectures by introducing (1) \textbf{Positional shortcuts}, where a 4×4 white patch is inserted at different spatial locations for different classes, and (2) \textbf{Intensity-based shortcuts}, where the pixel intensity in a specific region of the image is altered in a way that correlates with the class labels. Subsequently, the trained models are assessed for their reliance on shortcuts by evaluating their performance on both the shortcut-embedded and standard test sets. A model that effectively learns meaningful features should exhibit minimal performance variation across the two test sets, whereas a model heavily influenced by shortcuts will demonstrate a marked decline in accuracy when evaluated on the standard test set. Higher values for accuracy difference indicates greater susceptibility to shortcut learning, while a lower value suggests better generalization across the two test conditions.
\vspace{-1em}
\begin{figure}[h]
    \centering
    \includegraphics[width=1\textwidth]{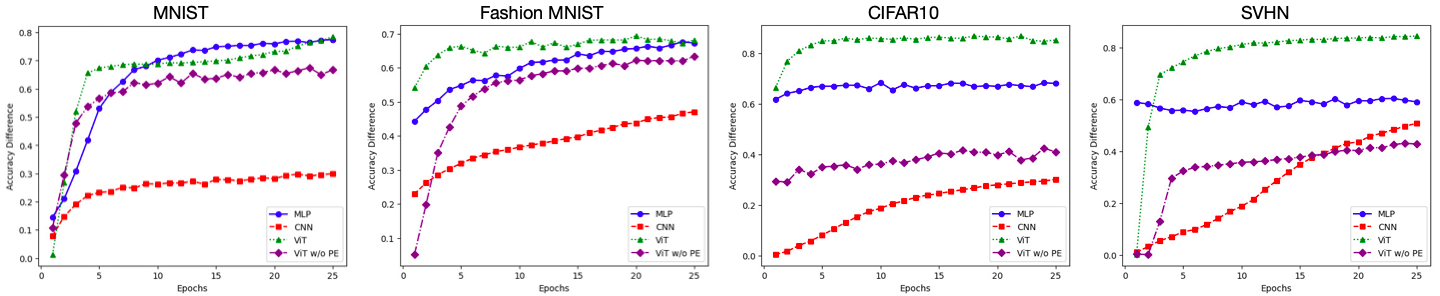}
    \caption{Accuracy Difference across architectures for positional shortcuts.}
    \label{fig:accuracy_diff_1}
    \vspace{-0.5em}
\end{figure}
\vspace{-0.5em}

In Figure~\ref{fig:accuracy_diff_1}, we observe a consistent trend across all datasets for the first type of shortcuts—the \textbf{positional shortcuts}. While accuracy on the shortcut test set is nearly perfect, accuracy on the normal test set is significantly degraded. Among the architectures, ViTs with positional encodings exhibit the highest susceptibility to shortcut learning, with the largest values for both Accuracy Difference and Loss Difference. This may be attributed to the presence of positional encodings in ViTs, which inherently reveal the spatial locations of the introduced shortcuts, making them easily learnable by the model. However, the results on ViTs without positional encodings are comparably better, with performance much closer to CNNs. On the other hand, CNNs demonstrate the best resistance to shortcuts, showing the smallest differences in accuracy and loss across the two test sets. MLPs display intermediate levels of shortcut susceptibility, performing better than ViTs but worse than CNNs, likely due to their lack of strong spatial priors.
\vspace{-1em}
\begin{figure*}[h]
    \centering
    \includegraphics[width=1\textwidth]{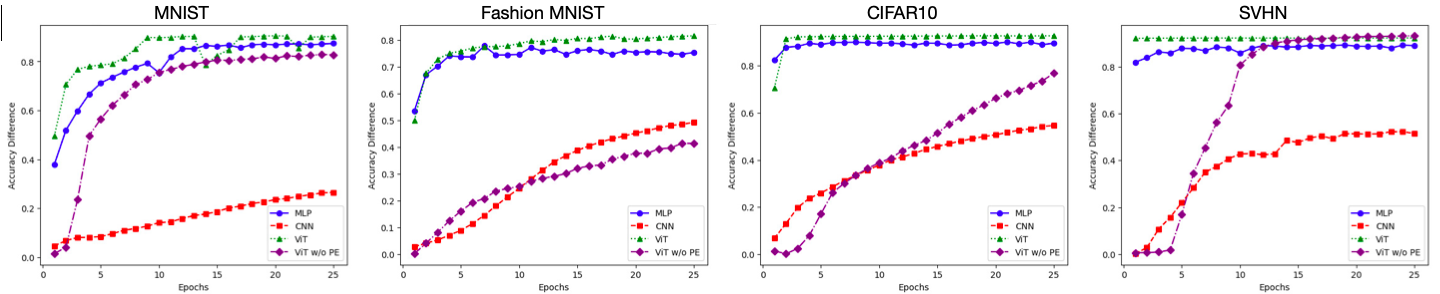}
    \caption{Accuracy Difference across architectures for intensity-based shortcuts.}
    \label{fig:accuracy_diff_2}
    \vspace{-1em}
\end{figure*}
\vspace{-1em}

In Figure~\ref{fig:accuracy_diff_2}, we observe a similar trend as earlier across all datasets for the second type of shortcuts—the \textbf{intensity-based shortcuts}. While accuracy on the shortcut test set remains high, accuracy on the normal test set is significantly lower, indicating that models leverage the intensity-based cues for classification rather than learning meaningful features.  Across architectures, CNNs again demonstrate the best resistance to shortcut learning, showing the smallest degradation in accuracy. However, unlike the positional shortcut case, MLPs now perform the worst, exhibiting the largest performance drop across datasets. For digit-based datasets like MNIST and SVHN, ViTs without positional encodings also struggle, performing similarly to MLPs. In contrast, on real-world datasets like CIFAR-10 and Fashion-MNIST, ViTs without positional encodings perform comparably to CNNs, indicating that they are less susceptible to intensity-based shortcuts in these settings.
\vspace{-1em}
\begin{figure}[h]
    \centering
    \includegraphics[width=1\textwidth]{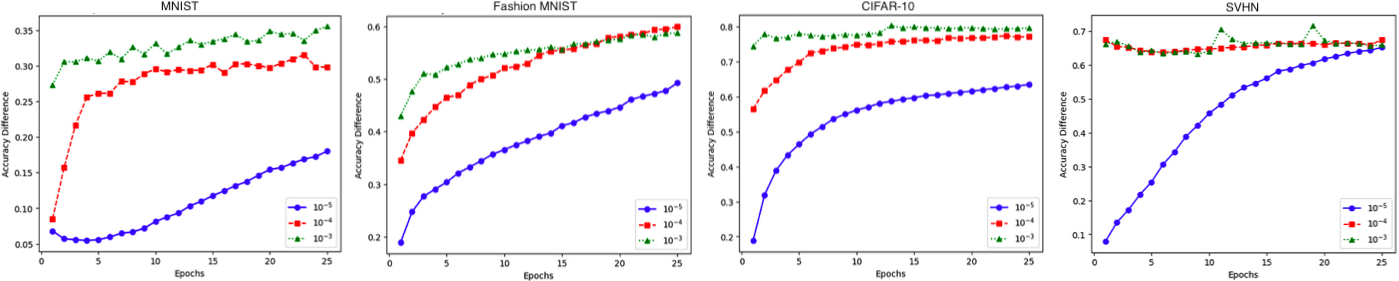}
    \caption{Accuracy Difference across different learning rates ($10^{-5}$, $10^{-4}$, $10^{-3}$).}
    \label{fig:accuracy_diff_lr}
    \vspace{-1em}
\end{figure}

Further, when trained with a large learning rate, models converge faster but tend to latch onto easily learnable patterns, including shortcuts, leading to higher Accuracy and Loss Difference values. In contrast, models trained with a small learning rate exhibit a more gradual learning process, favoring the acquisition of meaningful, class-relevant features over shortcut-based cues as shown in Figure~\ref{fig:accuracy_diff_lr}.
\vspace{-1em}

\begin{figure}[h]
    \centering
    \includegraphics[width=1\textwidth]{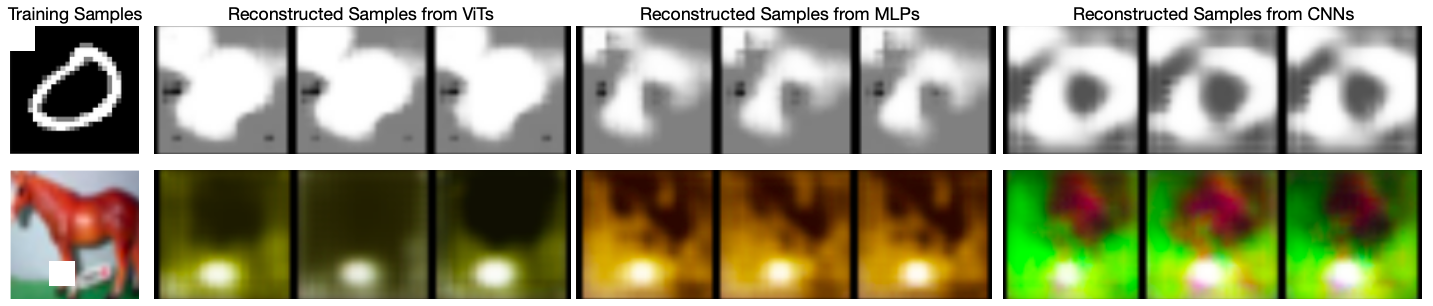}
    \caption{Reconstructed training samples as perceived by the classifier.}
    \label{fig:reconstruction_samples}
\end{figure}
\vspace{-1em}
Figure~\ref{fig:reconstruction_samples} presents a qualitative comparison of the perceived training images. The first column contains actual training samples with embedded shortcuts, while the subsequent columns show the reconstructed images for ViTs, MLPs, and CNNs respectively. We observe that CNNs tend to reconstruct meaningful features with spatial coherence, while MLPs exhibit moderate shortcut reliance, capturing a mix of class-relevant and shortcut features. ViTs, on the other hand, reconstruct highly distorted samples dominated by shortcut-based artifacts, suggesting a stronger reliance on positional correlations rather than intrinsic image structures.

\section{Conclusion}
In this paper, we evaluated the shortcut learning susceptibility of different vision classifier architectures across four benchmark datasets by introducing artificial shortcuts into the training data.

\bibliography{iclr2025_conference}
\bibliographystyle{iclr2025_conference}

\end{document}